\documentclass[letterpaper]{article} 
\usepackage{aaai2026}  
\usepackage{times}  
\usepackage{helvet}  
\usepackage{courier}  
\usepackage[hyphens]{url}  
\usepackage{graphicx} 
\usepackage{natbib}  
\usepackage{caption} 
\usepackage{amsmath, amssymb, amsfonts}
\usepackage{bm}
\usepackage{booktabs}
\usepackage{multirow}
\usepackage[table]{xcolor}
\definecolor{Gray}{gray}{0.9}

\DeclareMathOperator*{\argmax}{arg\,max}
\DeclareMathOperator*{\argmin}{arg\,min}
\newcommand{\reals}{\mathbb{R}}
\newcommand{\norm}[1]{\left\lVert#1\right\rVert}

\usepackage{algorithm}
\usepackage{algorithmic}

\title{Mosaic Pruning: A Hierarchical Framework for Generalizable Pruning of Mixture-of-Experts Models}

\author{
    Wentao Hu\textsuperscript{\rm 1, \rm 2}\equalcontrib,
    Mingkuan Zhao\textsuperscript{\rm 1}\equalcontrib,
    Shuangyong Song\textsuperscript{\rm 2},
    Xiaoyan Zhu\textsuperscript{\rm 1},
    Xin Lai\textsuperscript{\rm 1},\\
    Jiayin Wang\textsuperscript{\rm 1}\thanks{Corresponding author.}
}
\affiliations{
    \textsuperscript{\rm 1}Xi'an Jiaotong University\\
    \textsuperscript{\rm 2}Institute of Artificial Intelligence (TeleAI), China Telecom\\
    wentao\_hu@stu.xjtu.edu.cn, mingkuanzhao@stu.xjtu.edu.cn, songshy@chinatelecom.cn, zhu.xy@xjtu.edu.cn, laixin@xjtu.edu.cn, wangjiayin@mail.xjtu.edu.cn
}

\begin{document}

\maketitle

\begin{abstract}
Sparse Mixture-of-Experts (SMoE) architectures have enabled a new frontier in scaling Large Language Models (LLMs), offering superior performance by activating only a fraction of their total parameters during inference. However, their practical deployment is severely hampered by substantial static memory overhead, as all experts must be loaded into memory. Existing post-training pruning methods, while reducing model size, often derive their pruning criteria from a single, general-purpose corpus. This leads to a critical limitation: a catastrophic performance degradation when the pruned model is applied to other domains, necessitating a costly re-pruning for each new domain. To address this generalization gap, we introduce Mosaic Pruning (MoP). The core idea of MoP is to construct a functionally comprehensive set of experts through a structured ``cluster-then-select" process. This process leverages a similarity metric that captures expert performance across different task domains to functionally cluster the experts, and subsequently selects the most representative expert from each cluster based on our proposed Activation Variability Score. Unlike methods that optimize for a single corpus, our proposed Mosaic Pruning ensures that the pruned model retains a functionally complementary set of experts, much like the tiles of a mosaic that together form a complete picture of the original model's capabilities, enabling it to handle diverse downstream tasks.Extensive experiments on various MoE models demonstrate the superiority of our approach. MoP significantly outperforms prior work, achieving a 7.24\% gain on general tasks and 8.92\% on specialized tasks like math reasoning and code generation.
\end{abstract}

\begin{links}
    \link{Code}{https://github.com/Saul-James/MoP}
\end{links}

\section{Introduction}
\label{sec:introduction}

Large Language Models (LLMs) have recently demonstrated remarkable capabilities in complex reasoning and generation tasks~\citep{openai2024gpt4technicalreport,touvron2023llama2openfoundation,xiong-etal-2024-dual}. To mitigate their computational requirements, the Mixture-of-Experts (MoE) architecture has been widely adopted~\citep{jiang2024mixtralexperts}. MoE models, such as Mixtral 8x7B, activate only a portion of their parameters during inference, enabling them to surpass the performance of larger dense models like Llama 2 70B while maintaining a smaller count of active parameters~\citep{jiang2024mixtralexperts, touvron2023llama2openfoundation}.

Despite this efficiency, MoE models are constrained by a critical deployment challenge: immense static memory overhead. For instance, deploying Mixtral 8x7B requires over 80GB of GPU memory~\citep{jiang2024mixtralexperts}. More importantly, due to training dynamics and representation learning differences, significant redundancy exists among experts in MoE models~\citep{chi2022representation, yu2022diversifying}, with some experts being functionally similar or contributing minimally to most tasks.

Leveraging the similarity between experts, post-training pruning has emerged as a promising approach for compressing MoE models. The state-of-the-art pruning method, Enumeration Pruning~\citep{lu2024not}, utilizes a general-purpose calibration dataset (e.g., C4 and WikiText) to identify and remove experts that contribute least to token reconstruction loss. However, this method reveals a critical limitation: the inability to generalize to specialized downstream tasks. A model pruned on a general-purpose corpus suffers a catastrophic performance drop when directly applied to domain-specific tasks such as mathematical reasoning or code generation, a phenomenon we term functional collapse. We argue that this collapse occurs because such methods inherently favor retaining generalist experts that make moderate contributions across common data patterns, while discarding specialist experts that possess critical domain-specific expertise but exhibit less prominent activation on general datasets. This necessitates re-pruning with a new calibration dataset for each new domain, a process that is not only impractical but also fundamentally undermines the model's applicability. Given that different experts in MoE specialize in distinct knowledge domains, we argue that a model's generalization capability fundamentally depends on the breadth of its functional diversity.

To address these challenges, we first propose a pruning strategy that actively preserves expert diversity,which we term Global Variability-aware Pruning (GVP). This method calculates an Activation Variability Score ($S_{\text{var}}$), constructed based on the Kullback-Leibler (KL) divergence~\citep{kullback1951information}, for each expert and prioritizes retaining those with the highest scores. While our experiments confirm that GVP is superior to Enumeration Pruning, relying solely on the Activation Variability Score ($S_{\text{var}}$) may lead to the retention of two functionally similar experts that both possess high diversity scores, which still results in redundancy among the experts. To overcome this limitation, we propose a hierarchical pruning framework aimed at preserving functional diversity,which we call Mosaic Pruning (MoP). Our core idea is twofold: first, we use a composite metric, primarily driven by a Task Performance Similarity matrix ($S_{\text{perf}}$)~\citep{spearman1987proof}, to assess functional similarity by analyzing expert performance across different task domains; second, within each functionally defined group, we select the most critical representative based on the Activation Variability Score proposed in GVP. This strategy ensures that no essential functional expertise is entirely eliminated, thus providing a strong inductive bias toward generalization across diverse tasks. Our extensive experiments on models such as Mixtral-8x7B~\citep{jiang2024mixtralexperts} and Qwen1.5-MoE-A2.7B~\citep{team2024qwen2} demonstrate the superiority of our method.

Our contributions are summarized as follows:
\begin{itemize}
    \item We propose a novel diversity metric, the Activation Variability Score, to effectively identify experts with specialized functionalities.
    \item We introduce a novel pruning framework that first groups experts based on their task performance similarity, and then prunes within each group to select a functionally complementary set of specialists based on our diversity metric.
    \item We validate the superiority of our proposed method through extensive experiments, achieving an average performance improvement of 7.24\% on general tasks, and a more significant 8.92\% on specialized tasks such as mathematical reasoning and code generation.
\end{itemize}

\section{Related Works}
\label{sec:related_works}

\subsection{Mixture-of-Experts Models}
The Mixture-of-Experts (MoE) paradigm, first introduced by ~\citep{jacobs1991adaptive}, involves a collection of distinct sub-networks, or ``experts," coordinated by a trainable gating network that selectively routes inputs. This modular architecture was successfully integrated into modern deep learning with its application to Recurrent Neural Networks~\citep{shazeer2017outrageously} and later scaled up within large Transformer-based models, most notably in the encoder-decoder architecture of GShard~\citep{lepikhin2020gshard}. With the recent ascendancy of decoder-only architectures like the GPT series ~\citep{mann2020language, touvron2023llama, touvron2023llama2openfoundation}, there has been a surge in the development of powerful open-source MoE-based LLMs, such as Mixtral ~\citep{jiang2024mixtralexperts}, Qwen-MoE ~\citep{team2024qwen2}, and  DeepSpeed-MoE ~\citep{rajbhandari2022deepspeed}, and the Tele-FLM series~\citep{he2024telechattechnicalreport, li2024teleflmtechnicalreport, li202452b1tlessonslearned, 10884554, wang-etal-2024-telechat, wang2025technicalreporttelechat2telechat25}. A key advantage of MoE is its ability to decouple the model's total parameter count from its computational cost per token, enabling a massive increase in model capacity without a proportional rise in inference latency ~\citep{fedus2022switch}. However, this efficiency comes at the cost of significant memory overhead, as all experts must be loaded into memory, which motivates research into compressing these models.

\subsection{Model Pruning}
Model pruning aims to reduce the complexity of neural networks by removing redundant parameters, thereby enhancing efficiency. Pruning techniques are broadly categorized into two types: unstructured and structured pruning.

\textbf{Unstructured Pruning}, such as magnitude pruning ~\citep{han2015deep}, removes individual weights based on certain criteria. While it can achieve high sparsity levels, its practical application in LLMs has been challenging. Recent innovations like SparseGPT ~\citep{frantar2023sparsegpt} and Wanda ~\citep{sun2023simple} have revisited this concept for LLMs, developing sophisticated post-training methods that prune weights based on magnitude and activation statistics without requiring retraining. Other works reduce overhead via sparse attention~\citep{zhao2025makingheadcountsparse}. However, the irregular sparsity patterns produced by these methods often necessitate specialized hardware or software support to realize actual inference speedups.

\textbf{Structured Pruning}, in contrast, removes entire structural units of the network, such as neurons, attention heads, or, in the context of MoE models, entire experts. This form of pruning is inherently hardware-friendly, as it results in smaller, dense models that can be efficiently executed on standard hardware like GPUs. For MoE models, expert-level pruning is a particularly compelling form of structured pruning. Early works in this area often focused on task-specific scenarios, such as machine translation, where experts specializing in certain languages could be discarded ~\citep{kim2021scalable}. More recent approaches have focused on task-agnostic, post-training pruning for large-scale MoE LLMs. A prominent example is the reconstruction-based pruning method proposed by ~\citep{lu2024not}. This approach systematically enumerates expert combinations to identify the subset that minimizes token reconstruction loss on a general-purpose corpus. While effective at preserving performance on the calibration domain, our work identifies a critical limitation in this approach—a failure to generalize to specialized downstream tasks—which directly motivates our development of a diversity-aware pruning framework.

\section{Methodology}
\label{sec:methodology}

\begin{figure}[t]
    
    \centering

    \includegraphics[width=1.0\columnwidth]{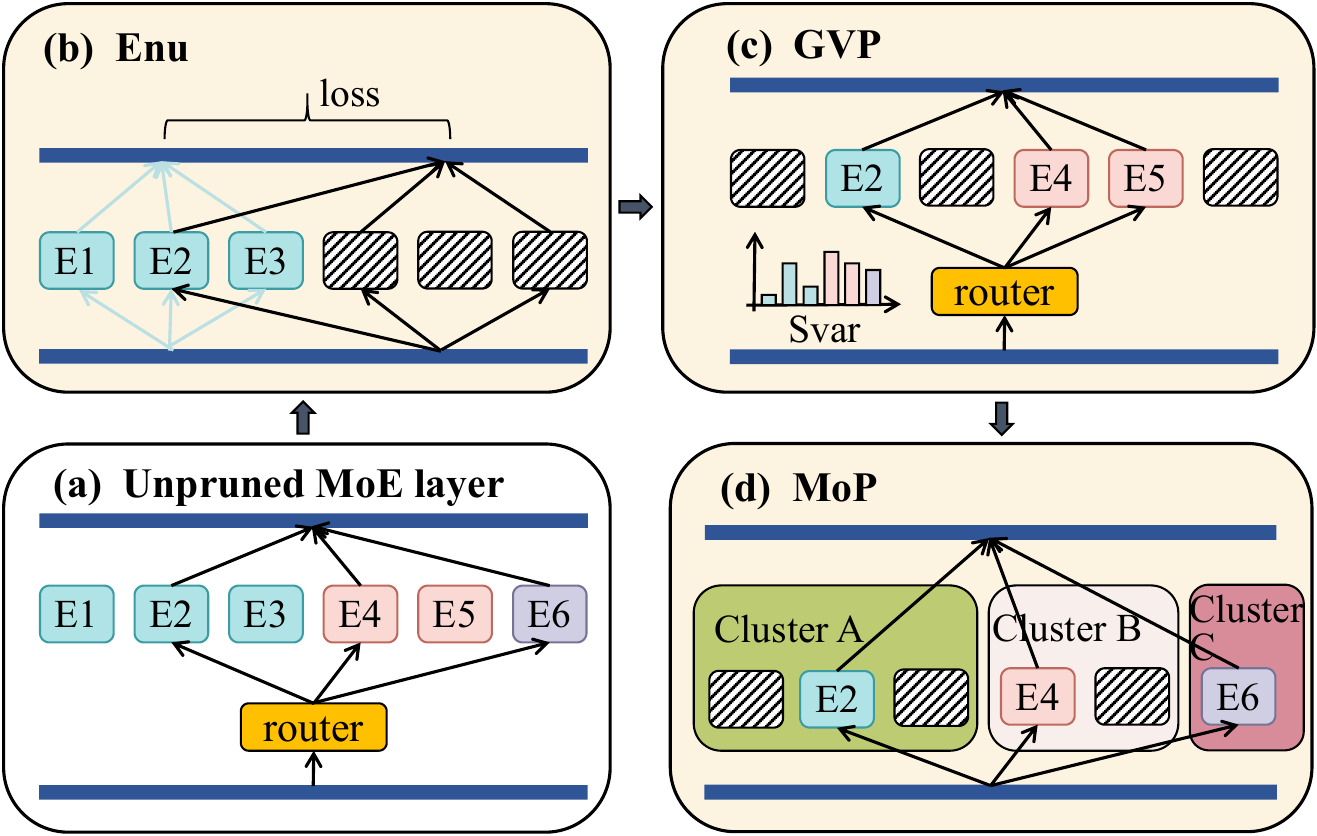}

    \caption{A conceptual illustration of the expert pruning strategies.
        (a) Unpruned MoE Layer: The original experts, with colors indicating different functional specializations.
        (b) Enu: Retains a functionally homogeneous set of experts (E1, E2, E3) by minimizing reconstruction loss.
        (c) GVP: Supplements core experts with globally selected specialists (E4, E5), improving diversity but risking functional overlap.
        (d) MoP: Clusters experts by functional similarity and selects a representative from each cluster, ensuring a final expert set that is both specialized and complementary.}

    \label{pic2}
\end{figure}

In this section, we present the details of our proposed hierarchical pruning framework. We introduce our contributions in a progressive sequence: first, we detail an improved strategy that incorporates global expert variability, which we term Global Variability-aware Pruning (GVP) (Section \ref{sec:global_var_pruning}). Building upon the insights and limitations of GVP, we then present our ultimate solution: a Hierarchical framework that combines functional clustering and diversity-aware selection, which we call Mosaic Pruning (MoP), designed to achieve an optimal balance between performance and diversity (Section \ref{sec:hierarchical_pruning}). Figure \ref{pic2} provides a conceptual overview of the pruning methods discussed in this paper.

\subsection{Enumeration Pruning (Enu)}
\label{sec:recon_pruning}

The most intuitive strategy for expert pruning is to quantify and minimize the perturbation to the model's output upon expert removal. This method (Enumeration Pruning), which forms the basis of our baseline, follows the core idea of ~\citep{lu2024not} by searching a general-purpose calibration dataset (e.g., C4) to find the optimal expert retention scheme. It aims to find the subset of $r$ experts, $\mathcal{E}_{\text{kept}}$, that minimizes the output reconstruction loss, an optimization problem formally expressed as:
\begin{equation}
\mathcal{E}^*_{\text{kept}} = \argmin_{\mathcal{E}_{\text{kept}} \subset \mathcal{E}, |\mathcal{E}_{\text{kept}}| = r} \mathcal{L}_{\text{recon}}(\mathcal{D}_{\text{cache}}, \mathcal{E}_{\text{kept}}),
\end{equation}
where the total reconstruction loss $\mathcal{L}_{\text{recon}}$ is calculated over a cached dataset $\mathcal{D}_{\text{cache}}$—a pre-computed set of input hidden states and their corresponding original layer outputs. The loss quantifies the cumulative difference between the outputs of the original, unpruned MoE layer and the pruned layer where only the experts in $\mathcal{E}_{\text{kept}}$ are active.
In our design, for models with a small number of experts, such as Mixtral 8x7B ($n=8$), we use enumeration to precisely solve this optimization problem. However, for models with a large number of experts, like Qwen1.5-MoE-A2.7B ($n=60$), the number of combinations $\binom{60}{r}$ renders exhaustive enumeration infeasible. In this case, we employ a greedy strategy as an approximation, iteratively removing the single expert that contributes the least to the increase in reconstruction loss until the target number is reached. While Enumeration Pruning is conceptually simple, as mentioned in the Introduction, it tends to preserve functionally generalist experts at the expense of specialist experts crucial for downstream professional tasks.

\subsection{Global Variability-aware Pruning (GVP)}
\label{sec:global_var_pruning}
To address the issue of Enumeration Pruning erroneously removing specialist experts due to its over-reliance on reconstruction loss, we propose Global Variability-aware Pruning (GVP). The core idea of this method is to apply a unified variability metric from a global perspective to prioritize the selection of functional specialists. The core metric of this method is the Activation Variability Score ($S_{\text{var}}$), which quantifies an expert's functional specialization. 

For any expert $i$, we define its variability as the Kullback-Leibler (KL) divergence~\citep{kullback1951information} between its normalized activation distribution and the uniform distribution over all tokens. Let $N_{\text{total}}$ be the total number of tokens in the calibration dataset. The score is calculated as:
\begin{equation}
S_{\text{var}}(i)=\sum_{t=1}^{N_{\text{total}}}
   \frac{p_{t,i}}{Z_i}\,
   \log_2\!\bigl(\tfrac{p_{t,i}}{Z_i}\,N_{\text{total}}\bigr).
\end{equation}
Here, $p_{t,i}$ is the Softmax activation probability for expert $i$ on token $t$, $Z_i = \sum_{t=1}^{N_{\text{total}}} p_{t,i}$ is the normalization constant, $P_i = \{\frac{p_{t,i}}{Z_i}\}_{t=1}^{N_{\text{total}}}$ is the normalized activation distribution, and $U = \{\frac{1}{N_{\text{total}}}\}_{t=1}^{N_{\text{total}}}$ is the uniform distribution over tokens. A high $S_{\text{var}}$ score indicates a concentrated activation pattern on a specific subset of inputs, suggesting it is a functional specialist.

The GVP pruning strategy is as follows:
\paragraph{Retaining the General Experts.} We first apply the process from Section \ref{sec:recon_pruning} to select a core set of $m$ ($m < r$) general experts ($\mathcal{E}_{\text{general}}$), which are most critical for maintaining baseline performance on general data and providing general capabilities:
\begin{equation}
\mathcal{E}_{\text{general}} = \argmin_{\mathcal{S} \subset \mathcal{E}, |\mathcal{S}| = m} \mathcal{L}_{\text{recon}}(\mathcal{D}_{\text{cache}}, \mathcal{S}).
\end{equation}
This step aims to identify and retain a set of the most functionally general experts to ensure that the model's core general capabilities are not compromised. The remaining $n-m$ experts in the candidate pool, $\mathcal{E}_{\text{cand}} = \mathcal{E} \setminus \mathcal{E}_{\text{general}}$, to be pruned according to a diversity score in the following selection stage.

\paragraph{Selecting Diverse Experts via Global Variability.} From the candidate pool $\mathcal{E}_{\text{cand}}$, we apply the Activation Variability score $S_{\text{var}}$ in a global ranking to select the remaining $r-m$ most specialized experts, forming the diversity set $\mathcal{E}_{\text{div}}$:
\begin{equation}
\mathcal{E}_{\text{div}} = \argmax_{\mathcal{S} \subset \mathcal{E}_{\text{cand}}, |\mathcal{S}| = r-m} \sum_{i \in \mathcal{S}} S_{\text{var}}(i).
\end{equation}
The final set of retained experts for the layer is $\mathcal{E}_{\text{final}}^{\text{GVP}} = \mathcal{E}_{\text{general}} \cup \mathcal{E}_{\text{div}}$. Although GVP is superior to the enumeration reconstruction loss strategy, its global ranking in the second stage has a critical flaw: it may retain two functionally similar yet highly specialized experts, which still results in redundancy within the final expert set.

\begin{figure*}[t]
    
    \centering

    \includegraphics[width=0.9\textwidth]{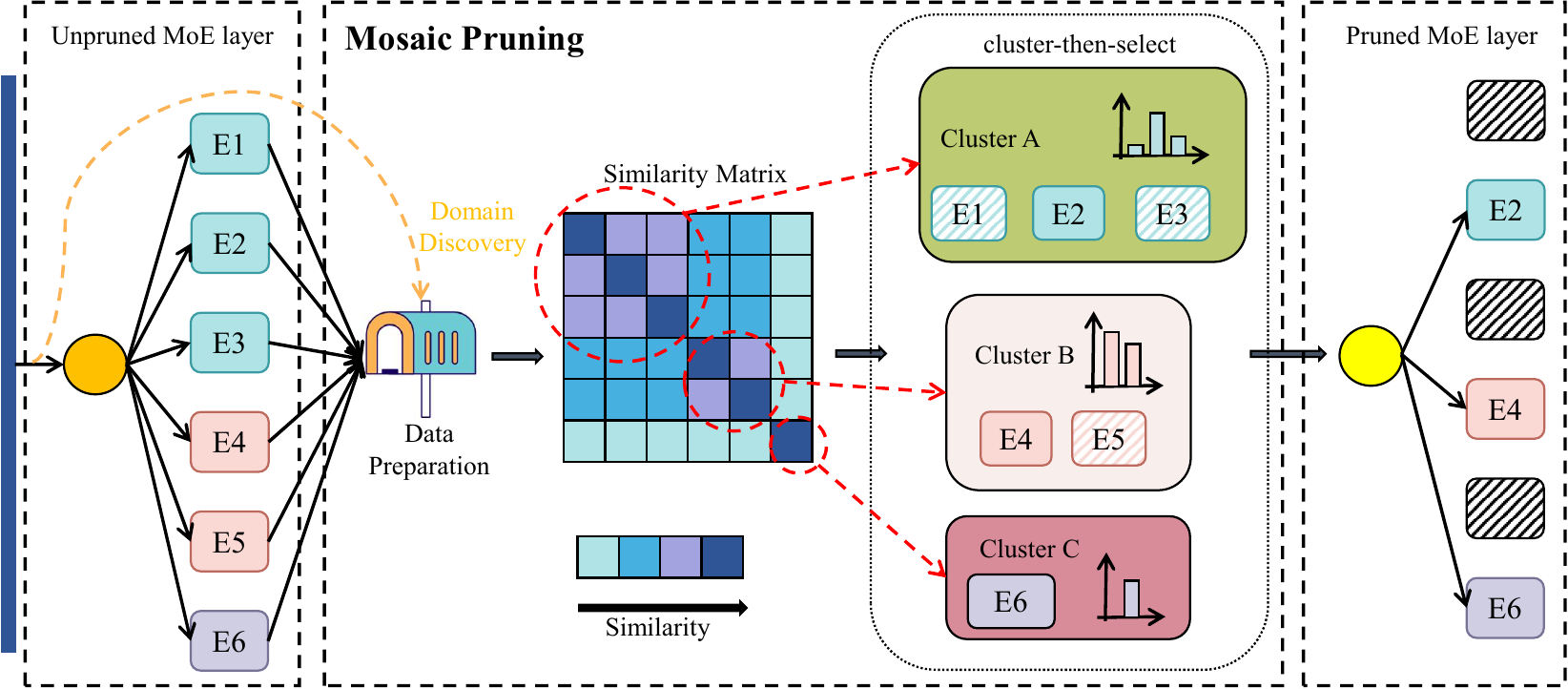}

    \caption{The workflow of the Mosaic Pruning (MoP) framework. First, the calibration data is partitioned into distinct functional domains. Subsequently, a similarity matrix between experts is constructed based on their performance profiles across these domains. This matrix is used to cluster experts with high similarity into the same group. Finally, experts are selected within each cluster based on their Activation Variability Score.}

    \label{pic1}
\end{figure*}

\subsection{Mosaic Pruning (MoP)}
\label{sec:hierarchical_pruning}
To fundamentally address the inherent flaws of GVP, we propose \textbf{Mosaic Pruning (MoP)}, a hierarchical framework that systematically preserves expert functional diversity. MoP shares the same first stage as GVP, retaining a shared expert set to secure baseline performance. Its core innovation lies in its second stage, which replaces GVP's simple global ranking with a structured ``cluster-then-select" process based on functional similarity, as illustrated in Figure \ref{pic1}.

\paragraph{Data-Driven Domain Discovery.}
\label{sec:Data-Driven Domain Discovery}

To evaluate expert performance across different functional domains, MoP utilizes a specially constructed mixed-diversity calibration dataset. This multi-domain design is crucial for enabling the automatic discovery of latent functional domains that align with expert specializations. The detailed construction process for this dataset is provided in Appendix A.
We apply K-Means~\citep{mcqueen1967some} clustering to the input hidden states of all tokens, $\{\mathbf{x}_t\}_{t=1}^{N_{\text{total}}}$. We posit that because experts exhibit domain-specific behaviors, and our mixed-diversity dataset is designed to reveal clear functional boundaries between different knowledge domains, K-Means is well-suited for capturing these domain-driven functional groupings by partitioning the token embeddings into semantically coherent domains. The objective is to find a partition $\mathcal{C} = \{C_1, C_2, \dots, C_K\}$ that minimizes the Within-Cluster Sum of Squares (WCSS):
\begin{equation}
\min_{\mathcal{C}} \sum_{k=1}^{K} \sum_{\mathbf{x}_t \in C_k} \norm{\mathbf{x}_t - \bm{\mu}_k}_2^2,
\end{equation}
where $\bm{\mu}_k = \frac{1}{|C_k|}\sum_{\mathbf{x}_t \in C_k} \mathbf{x}_t$ is the centroid of the $k$-th cluster.
We set the number of clusters $K = r-m$. This design choice creates a one-to-one mapping between the discovered functional clusters and the experts to be selected, providing a strong inductive bias towards preserving each distinct functional capability while minimizing redundancy. This process assigns a domain label to each token, defining a mapping function $\text{domain}: \{1, \dots, N_{\text{total}}\} \to \{1, \dots, K\}$.

\begin{table*}[t] 
\centering
\small 

\begin{tabular}{clccccccccc}
\toprule
\textbf{Model} & \textbf{Method} & \textbf{Experts} & \textbf{ARC-c} & \textbf{ARC-e} & \textbf{BoolQ} & \textbf{HellaSwag} & \textbf{MMLU} & \textbf{OBQA} & \textbf{WinoGrande} & \textbf{Average} \\
\midrule

\multirow{9}{*}{\textbf{Mixtral}} & \textbf{None} & 8 & 80.21 & 87.29 & 81.47 & 75.91 & 67.43 & 83.20 & 53.26 & 75.54 \\
\cmidrule(lr){2-11}
 & \multirow{2}{*}{\textbf{Random}} & 6 & 69.97 & 78.11 & 83.18 & 68.64 & 53.44 & 74.20 & 53.15 & 68.67 \\
 &                          & 4 & 44.54 & 47.22 & 64.86 & 59.83 & 48.21 & 67.20 & 55.72 & 55.37 \\
\cmidrule(lr){2-11}
 & \multirow{2}{*}{\textbf{Frequency}} & 6 & 70.90 & 78.15 & 78.28 & 56.85 & 58.47 & 78.20 & 58.25 & 68.44 \\
 &                            & 4 & 50.26 & 64.56 & 67.46 & 37.50 & 46.76 & 63.20 & 50.27 & 54.29 \\
\cmidrule(lr){2-11}
 & \multirow{2}{*}{\textbf{Enu}} & 6 & 73.04 & 83.16 & 84.89 & 67.49 & 61.14 & 81.00 & 58.41 & 72.73 \\
 &                                  & 4 & 59.59 & 74.41 & 83.12 & 52.02 & 49.45 & 63.80 & 54.46 & 62.41 \\
\cmidrule(lr){2-11}
 & \multirow{2}{*}{\textbf{MoP (Ours)}} & 6 & 75.43 & 82.37 & 86.51 & 74.28 & 63.50 & 80.20 & 55.80 & \textbf{74.01} \\
 &                              & 4 & 69.62 & 79.84 & 83.33 & 68.85 & 54.26 & 72.00 & 54.85 & \textbf{68.96} \\
\midrule

\multirow{9}{*}{\textbf{Qwen}} & \textbf{None} & 60 & 54.18 & 64.56 & 78.90 & 58.99 & 37.44 & 61.00 & 48.38 & 57.64 \\
\cmidrule(lr){2-11}
 & \multirow{2}{*}{\textbf{Random}} & 50 & 47.44 & 58.75 & 39.14 & 68.25 & 54.38 & 66.40 & 31.41 & 52.25 \\
 &                          & 40 & 38.57 & 50.63 & 1.92  & 48.20 & 36.37 & 38.20 & 40.88 & 36.40 \\
\cmidrule(lr){2-11}
 & \multirow{2}{*}{\textbf{Frequency}} & 50 & 41.98 & 57.99 & 23.82 & 36.10 & 41.80 & 50.80 & 46.17 & 42.67 \\
 &                            & 40 & 30.29 & 36.19 & 9.05  & 59.18 & 47.81 & 59.60 & 49.32 & 41.63 \\
\cmidrule(lr){2-11}
 & \multirow{2}{*}{\textbf{Enu}} & 50 & 54.61 & 66.16 & 33.03 & 62.89 & 53.69 & 65.20 & 50.19 & 55.11 \\
 &                                  & 40 & 30.63 & 35.65 & 22.05 & 42.36 & 43.49 & 49.80 & 46.25 & 38.60 \\
\cmidrule(lr){2-11}
 & \multirow{2}{*}{\textbf{MoP (Ours)}} & 50 & 60.70 & 75.04 & 38.10 & 67.28 & 55.60 & 66.80 & 50.19 & \textbf{59.10} \\
 &                              & 40 & 44.97 & 59.76 & 68.56 & 56.49 & 48.44 & 59.40 & 52.57 & \textbf{55.74} \\
\bottomrule
\end{tabular}
\caption{Zero-shot performance of MoP compared with baseline pruning methods. MoP is compared with Enumeration(Enu), Random, and Frequency Pruning. Experts indicates the number of experts retained per layer. Best average scores are in \textbf{bold}.}
\label{tab:main_results}
\end{table*}

\paragraph{Similarity Matrix Construction.}
For each pair of experts $(i, j)$ in the candidate pool $\mathcal{E}_{\text{cand}}$, we compute a similarity score $S_{\text{comp}}(i, j)$. In this framework, an expert’s functional similarity is measured by its performance across diverse task types. We choose Spearman's rank correlation coefficient as the core metric for task performance similarity because of three key advantages: First, as a nonparametric method, it requires no assumptions about the data distribution; second, its rank-based nature makes it inherently robust to outliers; and most importantly, it effectively captures the monotonic relationship of expert performance across different task domains, which aligns perfectly with our goal of identifying experts with the same functional profile. Therefore, this score is defined solely by the task performance similarity $S_{\text{perf}}$:
\begin{equation}
S_{\text{comp}}(i, j) = S_{\text{perf}}(i, j).
\end{equation}
To compute $S_{\text{perf}}$, we first construct a performance vector $\mathbf{v}_i^{\text{perf}} \in \reals^K$ for expert $i$. Its $k$-th element, $v_{i,k}^{\text{perf}}$, is the expert's average
reconstruction error in domain $k$ (smaller values mean better performance).
Because Spearman correlation is computed on ranks, this monotonic
definition preserves functional similarity even though lower numbers
indicate higher skill. Let $\mathcal{T}_k = \{t | \text{domain}(t)=k\}$ denote the set of token indices belonging to domain $k$. The error is then calculated as:
\begin{equation}
    v_{i,k}^{\text{perf}} = \frac{1}{|\mathcal{T}_k|} \sum_{t \in \mathcal{T}_k} \norm{E_i(\mathbf{x}_t) - \mathbf{z}_{\text{real}, t}}_2^2.
\end{equation}
Here, $E_i(\mathbf{x}_t)$ represents the output of the MoE layer only with expert $i$ when given the input token $\mathbf{x}_t$, effectively simulating a scenario where only this single expert is activated. This performance vector $\mathbf{v}_i^{\text{perf}}$ serves as a functional profile for expert $i$, characterizing its effectiveness across the discovered domains.
To measure the monotonic relationship, we compute the Spearman rank correlation coefficient. Let $\text{rank}(\cdot)$ be the rank transformation operator. For a given performance vector $\mathbf{v}_i^{\text{perf}} = [v_{i,1}, \dots, v_{i,K}]$, this operator sorts the elements and replaces each element with its ordinal rank (e.g., the smallest element is replaced by rank 1, the second smallest by rank 2, and so on). This converts the performance vector into a rank vector $\mathbf{R}_i = \text{rank}(\mathbf{v}_i^{\text{perf}})$. For clarity, we first define the sum of squared deviations for a centered rank vector as:
\begin{equation}
    SS_i = \sum_{k=1}^K (R_{ik} - \bar{R}_i)^2,
\end{equation}
where $\bar{R}_i$ is the mean of the elements in the rank vector $\mathbf{R}_i$. Based on this, the formula for the Spearman correlation coefficient can be concisely expressed as:
\begin{equation}
    \rho(\mathbf{R}_i, \mathbf{R}_j) = \frac{\sum_{k=1}^K (R_{ik} - \bar{R}_i)(R_{jk} - \bar{R}_j)}{\sqrt{SS_i \cdot SS_j}}.
\end{equation}
Finally, we normalize the $\rho$ value to the interval $[0, 1]$ to obtain the final performance similarity score:
\begin{equation}
    S_{\text{perf}}(i, j) = \frac{1}{2} \left( 1 + \rho(\mathbf{R}_i, \mathbf{R}_j) \right).
\end{equation}

\begin{table*}[t] 
\centering
\small 
\begin{tabular}{clccccccccc}
\toprule
\textbf{Model} & \textbf{Method} & \textbf{Experts} & \textbf{ARC-c} & \textbf{ARC-e} & \textbf{BoolQ} & \textbf{HellaSwag} & \textbf{MMLU} & \textbf{OBQA} & \textbf{WinoGrande} & \textbf{Average} \\
\midrule

\multirow{4}{*}{\textbf{Mixtral}} & \textbf{Enu(C4)} & 6 & 73.04 & 83.16 & 84.89 & 67.49 & 61.14 & 81.00 & 58.41 & 72.73 \\
 & \textbf{Enu(Mixed)} & 6 & 74.57 & 85.35 & 84.06 & 71.55 & 60.91 & 79.80 & 53.28 & 72.79 \\
 & \textbf{GVP} & 6 & 76.62 & 84.09 & 85.20 & 73.79 & 63.43 & 81.00 & 50.35 & 73.50 \\
 & \textbf{MoP (Ours)} & 6 & 75.43 & 82.37 & 86.51 & 74.28 & 63.50 & 80.20 & 55.80 & \textbf{74.01} \\
\midrule

\multirow{4}{*}{\textbf{Qwen}} & \textbf{Enu(C4)} & 50 & 54.61 & 66.16 & 33.03 & 62.89 & 53.69 & 65.20 & 50.19 & 55.11 \\
 & \textbf{Enu(Mixed)} & 50 & 41.81 & 49.45 & 15.84 & 66.27 & 54.84 & 68.20 & 49.88 & 49.47 \\
 & \textbf{GVP} & 50 & 54.52 & 65.57 & 33.42 & 63.83 & 54.56 & 67.40 & 49.64 & 55.56 \\
 & \textbf{MoP (Ours)} & 50 & 60.70 & 75.04 & 38.10 & 67.28 & 55.60 & 66.80 & 50.19 & \textbf{59.10} \\
\bottomrule
\end{tabular}
\caption{Zero-shot performance ablation study of different pruning methods (some evaluated on different calibration dataset). All methods retain the same number of experts. Best average scores are in \textbf{bold}.}
\label{tab:ablation}
\end{table*}

\paragraph{Agglomerative Hierarchical Clustering.} 
After obtaining the similarity matrix $S_{\text{comp}}$, we convert it into a distance matrix $\mathbf{D}$, where the distance between any two experts $i$ and $j$ is defined as $D(i,j) = 1 - S_{\text{comp}}(i, j)$. This distance matrix conceptually defines the space in which the clustering operates.

We then apply hierarchical clustering, a bottom-up approach that iteratively merges the closest pair of clusters. The ``closeness" of two clusters is determined by a linkage criterion. We employ Ward's linkage method~\citep{ward1963hierarchical}, which defines the cost of merging two clusters, $C_a$ and $C_b$, as the increase in the total Error Sum of Squares (ESS) that would result from their merger. This merge cost, denoted as $\Delta(\text{ESS})$, serves as the specific inter-cluster distance measure for Ward's method and is calculated as:
\begin{equation}
    \Delta(\text{ESS}(C_a, C_b)) = \frac{|C_a||C_b|}{|C_a|+|C_b|} \norm{\bm{\mu}_a - \bm{\mu}_b}_2^2.
\end{equation}
Here, $\bm{\mu}_a$ and $\bm{\mu}_b$ are the centroids of the clusters, defined as the mean of the performance vectors ($\mathbf{v}^{\text{perf}}$) of the experts within each cluster:
\begin{equation}
\bm{\mu}_a = \frac{1}{|C_a|} \sum_{i \in C_a} \mathbf{v}_i^{\text{perf}}.
\end{equation}
At each step, the algorithm merges the pair of clusters that minimizes this merge cost. This process continues until we are left with $r-m$ clusters, resulting in a partition of $\mathcal{E}_{\text{cand}}$, $\mathcal{G} = \{G_1, G_2, \dots, G_{r-m}\}$.

\paragraph{Intra-Cluster Representative Selection.}
The final step is to select a representative from each functional cluster. We use the Activation Variability Score ($S_{\text{var}}$) as the selection criterion to identify the most specialized expert within each group. For each cluster $G_k$, the representative expert $e_k^*$ is chosen as:
\begin{equation}
e_k^* = \argmax_{i \in G_k} S_{\text{var}}(i).
\end{equation}
The resulting set of representatives, $\mathcal{E}_{\text{div}}$, forms a group of experts that are both individually specialized and functionally complementary.

\section{Experiments}
\label{sec:experiments}

\subsection{Experimental Setup}
\label{sec:exp_setup}

\paragraph{Models and Benchmarks.}
\label{sec:models-benchmarks}
Our experiments are conducted on two representative open-source MoE models: Mixtral-8x7B-Instruct (47B parameters, 8 experts/layer) and Qwen1.5-MoE-A2.7B-Chat (14.3B parameters, 60 experts/layer). This significant difference in expert count allows us to evaluate the robustness and scalability of our method. We evaluate performance on a wide range of benchmarks. For general capability assessment, we use seven standard benchmarks: ARC-c/e~\citep{clark2018think}, BoolQ~\citep{clark2019boolq}, HellaSwag~\citep{zellers2019hellaswag}, MMLU~\citep{hendrycks2020measuring}, OpenBookQA~\citep{mihaylov2018can}, and WinoGrande~\citep{sakaguchi2021winogrande}. For specialized skill assessment, we use four benchmarks across two domains: mathematical reasoning (GSM8K~\citep{cobbe2021training}, MATH~\citep{hendrycks2021measuring}) and code generation (HumanEval~\citep{chen2021evaluating}, MBPP~\citep{austin2021program}). Finally, for expert diversity validation (Section \ref{sec:diversity_analysis}), we utilize test samples from six distinct domains: Math-GSM8K~\citep{cobbe2021training}, Code-HumanEval~\citep{chen2021evaluating}, Commonsense-PIQA~\citep{bisk2020piqa}, Summarization-XSum~\citep{narayan2018don}, Science-ARC-C~\citep{clark2018think}, and Logic-MNLI~\citep{williams2017broad}.

\paragraph{Implementation Details.} Our main baseline is Enumeration Pruning~\citep{lu2024not}. Following its original implementation, this method uses a calibration dataset sourced from the general-purpose C4 corpus~\citep{raffel2020exploring}. In contrast,to improve functional clustering,our GVP and MoP methods use a mixed-diversity calibration dataset as described in (Section\ref{sec:Data-Driven Domain Discovery}). All experiments were conducted on a server equipped with four NVIDIA A100 80G GPUs.

\subsection{Main Results}
\label{sec:main_results}

\paragraph{Performance of Mosaic Pruning on General Tasks.}
We compare MoP against three baseline methods: Enumeration Pruning~\citep{lu2024not}, Random Pruning (which randomly discards experts), and Frequency Pruning ~\citep{muzio2024seer} (which removes the least frequently activated experts in each layer). As shown in Table \ref{tab:main_results}, MoP consistently and significantly outperforms all baseline methods in all pruning rates and models, achieving an average performance improvement of 7.24\% over Enumeration Pruning. These results demonstrate that preserving functional diversity enables more robust model compression.

\paragraph{Inference Speed and Memory Usage.}
As a structured pruning method, MoP's advantages are directly reflected in deployment efficiency. We evaluated the inference speed and peak memory usage of the models before and after pruning. As shown in Table \ref{tab:efficiency}, removing entire expert modules resulted in significant memory savings. Concurrently, we observed a significant acceleration in inference, achieving an average speedup of 1.20×, due to the reduced number of experts and potential inter-device communication overhead.

\begin{table}[h!]
    \centering
    \begin{tabular}{llccc}
        \toprule
        \textbf{Model} & \textbf{Method} & \textbf{Experts} & \textbf{Mem} & \textbf{Speedup} \\
        \midrule
        
        \multirow{3}{*}{\textbf{Mixtral}} & \textbf{None} & 8 & 87.87 & 1.00x \\
        \cmidrule(lr){2-5} 
         & \multirow{2}{*}{\textbf{MoP}} & 6 & 66.87 & 1.22x \\
         & & 4 & 45.87 & 1.52x \\
         
        \midrule
        
        \multirow{3}{*}{\textbf{Qwen}} & \textbf{None} & 60 & 27.79 & 1.00x \\
        \cmidrule(lr){2-5}
         & \multirow{2}{*}{\textbf{MoP}} & 50 & 23.91 & 1.12x \\
         & & 40 & 20.04 & 1.36x \\
         
        \bottomrule
    \end{tabular}
    \caption{Inference speed and memory(GB) usage of MoP-pruned models compared to the original architectures.}
    \label{tab:efficiency}
\end{table}

\begin{figure*}[t]
    
    \centering

    \includegraphics[width=0.83\textwidth]{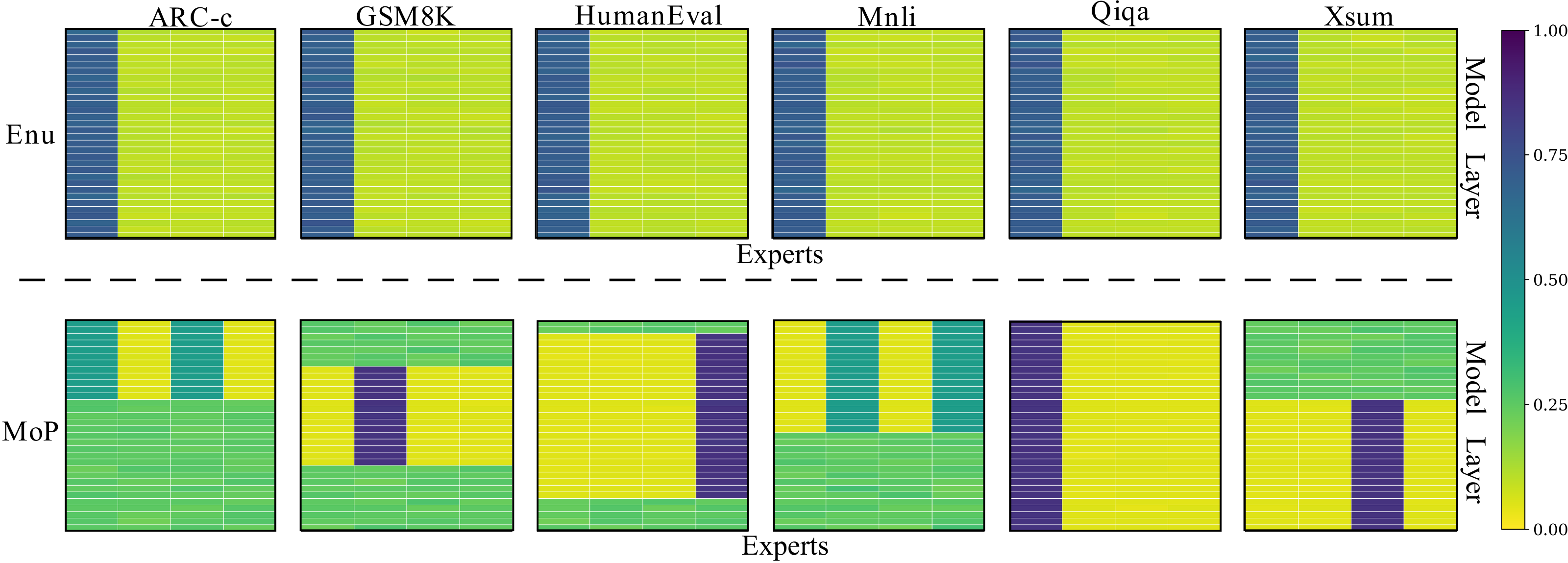}

    \caption{Heatmaps of expert activation weights across different domains for Mixtral-8x7B pruned to 4 experts. \textbf{(Top Row)}: The Enumeration Pruning leads to functional homogenization, with a few generalist experts dominating all tasks. \textbf{(Bottom Row)}: Our MoP method preserves domain specialization, with different experts activating for different tasks.}

    \label{div1}
\end{figure*}

\subsection{Ablation Study}
\label{sec:ablation}
To clarify the contribution of each component, we conducted ablation studies on Enumeration Pruning, GVP, and MoP. As shown in Table \ref{tab:ablation}, from Enumeration Pruning to GVP to MoP, performance shows progressive improvements. Since MoP and Enumeration Pruning used different calibration datasets, to ensure fair comparison, we re-experimented using the same mixed-diversity calibration dataset, comparing MoP with Enumeration Pruning (Mixed). The results show that even with the same calibration dataset, MoP still outperforms Enumeration Pruning (Mixed) across all benchmarks, demonstrating the superiority of our pruning framework.

\begin{table}[t]
    
    \centering
    
    \begin{tabular}{llccc}
        \toprule 
        
        \textbf{Model} & \textbf{Method} & \textbf{Experts} & \textbf{GSM8K} & \textbf{MATH} \\
        \midrule

        \multirow{4}{*}{\textbf{Mixtral}} & \multirow{2}{*}{\textbf{Enu}}  & 6  & 28.73 & 9.00  \\
                                          &                            & 4  & 22.37 & 3.80  \\
        
        \cmidrule(lr){2-5}
                                          & \multirow{2}{*}{\textbf{MoP}} & 6  & 51.33 & 9.20  \\
                                          &                            & 4  & 28.35 & 3.00  \\
        \midrule

        \multirow{4}{*}{\textbf{Qwen}}     & \multirow{2}{*}{\textbf{Enu}}  & 50 & 34.34 & 7.39  \\
                                          &                            & 40 & 16.76 & 5.40  \\
        \cmidrule(lr){2-5}
                                          & \multirow{2}{*}{\textbf{MoP}} & 50 & 35.63 & 11.20 \\
                                          &                            & 40 & 21.91 & 7.20  \\
                                          
        \bottomrule 
    \end{tabular}
    
    \caption{Zero-shot performance comparison on mathematical reasoning benchmarks at various pruning rates.}
    \label{tab:specialized_tasks_math}
\end{table}

\subsection{Validation of Expert Diversity and Specialization}
\label{sec:diversity_analysis}
\paragraph{Analysis of Expert Activation Diversity.}

To examine whether the pruned experts still cover a wide range of functional areas, we conducted activation pattern analysis on MoP and Enumeration Pruning. We used test samples from six different domains as described in (Section\ref{sec:models-benchmarks}), recorded the activation weights of the four experts pruned from Mixtral-8x7B across all 32 MoE layers, calculated average values to generate 32×4 matrices, and visualized them as heatmaps in Figure\ref{div1}. The results show that MoP maintains clear domain specialization, with different tasks activating different experts; while Enumeration Pruning leads to functional homogenization, where generalist experts dominate across all domains.

\paragraph{Performance Validation on Specialized Tasks.}

To quantitatively confirm the observations from our diversity analysis, we conducted a direct comparison between MoP and Enumeration Pruning methods on highly specialized downstream tasks. We evaluated the models in two distinct domains: mathematical reasoning and code generation. As shown in Table \ref{tab:specialized_tasks_math} and Table \ref{tab:specialized_tasks_code}, The results provide clear evidence of functional collapse in existing methods: the MoP-pruned model far outperforms Enumeration Pruning across the four specialized benchmarks, with an average performance improvement of 8.92\%. By preserving functional diversity and specialization, our method achieves a high compression rate while avoiding this functional collapse and maintaining the model's complex reasoning abilities in various specialized domains to the greatest extent possible.

\begin{table}[h!] 
    \centering 
    
    \begin{tabular}{llccc} 
        \toprule 
        
        \textbf{Model} & \textbf{Method} & \textbf{Experts} & \textbf{HumanEval} & \textbf{MBPP} \\
        \midrule

        \multirow{4}{*}{\textbf{Mixtral}} & \multirow{2}{*}{\textbf{Enu}}  & 6  & 82.31 & 14.00 \\
                                          &                            & 4  & 1.22  & 1.55  \\
        \cmidrule(lr){2-5} 
                                          & \multirow{2}{*}{\textbf{MoP}} & 6  & 83.54 & 15.95 \\
                                          &                            & 4  & 71.34 & 10.89 \\
        \midrule

        \multirow{4}{*}{\textbf{Qwen}}     & \multirow{2}{*}{\textbf{Enu}}  & 50 & 1.21  & 8.17  \\
                                          &                            & 40 & 1.83  & 1.95  \\
        \cmidrule(lr){2-5} 
                                          & \multirow{2}{*}{\textbf{MoP}} & 50 & 14.02 & 9.34  \\
                                          &                            & 40 & 3.05  & 8.17  \\
                                          
        \bottomrule 
    \end{tabular}
    \caption{Zero-shot performance comparison on code generation benchmarks at various pruning rates.}
    \label{tab:specialized_tasks_code} 
\end{table}

\section{Conclusion}

In this paper, we address the poor generalization of existing MoE pruning methods, which discard specialist experts, by proposing Mosaic Pruning (MoP). MoP is a hierarchical ``cluster-then-select" framework that preserves a functionally comprehensive set of experts. Extensive experiments show our method outperforms existing baselines, particularly on specialized tasks. Our work establishes a ``prune once, deploy efficiently across diverse tasks" paradigm, enhancing the practical applicability of large-scale MoE models.

\section{Acknowledgments}
This work was supported by the National Natural Science Foundation of China (Grant Nos. 62572389, 72293581, 72274152, 62402376). We also acknowledge the support from Xi'an Jiaotong University and the Institute of Artificial Intelligence of China Telecom (TeleAI).
\bibliography{aaai2026} 

\end{document}